\documentclass[lettersize,journal]{IEEEtran}
\usepackage{amsmath,amsfonts}
\usepackage{algorithmic}
\usepackage{algorithm}
\usepackage{array}
\usepackage[caption=false,font=normalsize,labelfont=sf,textfont=sf]{subfig}
\usepackage{textcomp}
\usepackage{stfloats}
\usepackage{url}
\usepackage{verbatim}
\usepackage{graphicx}
\usepackage{cite}
\usepackage{longtable}
\usepackage{booktabs}
\usepackage{multirow}
\usepackage{caption}
\begin{document}

\title{MedSAM-U: Uncertainty-Guided Auto Multi-Prompt Adaptation for Reliable MedSAM}

\author{Nan Zhou, Ke Zou, Kai Ren, Mengting Luo, Linchao He,\\ Meng Wang, Yidi Chen, Yi Zhang,  \IEEEmembership{Senior Member, IEEE}, Hu Chen, and Huazhu Fu, \IEEEmembership{Senior Member, IEEE}
\thanks{This work was supported by the Sichuan Science and Technology Program under Grant 2022JDJQ0045, and the Chengdu Key Research and development Support project under Grant 2024YF0500910SN }
\thanks{N. Zhou, K. Zou, K. Ren, M. Luo, L. He, and H. Chen are with the College of Computer Science, Sichuan University, Chengdu 610065, China.}
\thanks{Meng Wang is with the Centre for Innovation \& Precision Eye Health, Department of Ophthalmology, Yong Loo Lin School of Medicine, National University of Singapore, Singapore 119228.}
\thanks{Y. Chen is with the Department of Radiology, West China Hospital, Sichuan University, Chengdu 610065, China.}
\thanks{Y. Zhang is with the School of Cyber Science and Engineering and the Key Laboratory of Data Protection and Intelligent Management, Ministry of Education, Sichuan University, Chengdu 610065, China.}
\thanks{Huazhu Fu is with the Institute of High Performance Computing (IHPC), Agency for Science, Technology and Research (A*STAR), Singapore 138632. }
\thanks{N. Zhou and K. Zou, contributed equally to this work.}
\thanks{Hu Chen and Huazhu Fu are the co-corresponding authors (e-mail: huchen@scu.edu.cn, hzfu@ieee.org). }
}



\maketitle

\begin{abstract}
The Medical Segment Anything Model (MedSAM) has shown remarkable performance in medical image segmentation, drawing significant attention in the field. However, its sensitivity to varying prompt types and locations poses challenges. This paper addresses these challenges by focusing on the development of reliable prompts that enhance MedSAM's accuracy. We introduce MedSAM-U, an uncertainty-guided framework designed to automatically refine multi-prompt inputs for more reliable and precise medical image segmentation. Specifically, we first train a Multi-Prompt Adapter integrated with MedSAM, creating MPA-MedSAM, to adapt to diverse multi-prompt inputs. We then employ uncertainty-guided multi-prompt to effectively estimate the uncertainties associated with the prompts and their initial segmentation results. In particular, a novel uncertainty-guided prompts adaptation technique is then applied automatically to derive reliable prompts and their corresponding segmentation outcomes. We validate MedSAM-U using datasets from multiple modalities to train a universal image segmentation model. Compared to MedSAM, experimental results on five distinct modal datasets demonstrate that the proposed MedSAM-U achieves an average performance improvement of 1.7\% to 20.5\% across uncertainty-guided prompts.\end{abstract}


\begin{figure}[htbp]
\centering
\includegraphics[width=1\linewidth]{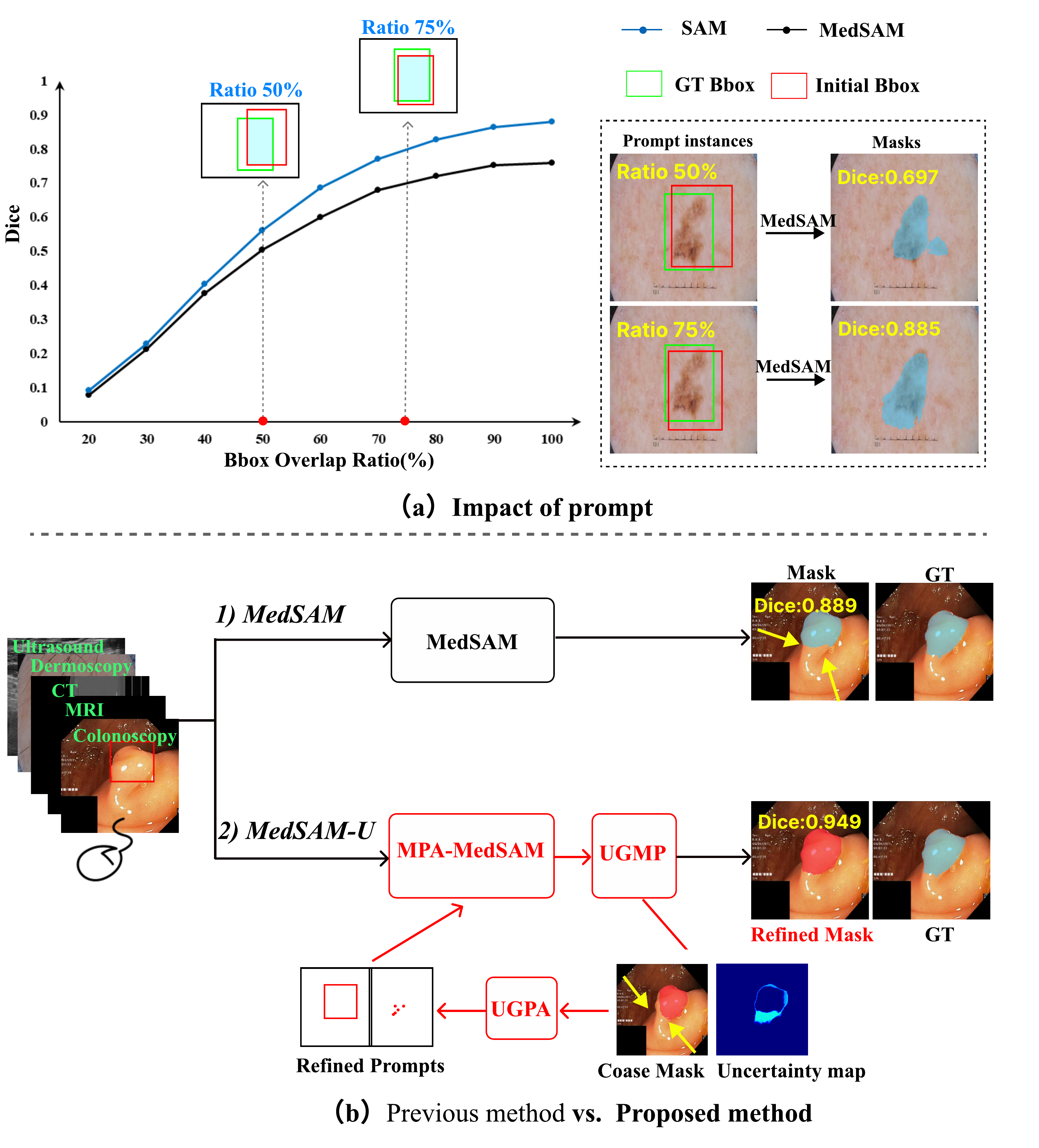}

\caption{ Comparison of Dice Results Under (a) Bboxes with different overlap ratio on SAM and MedSAM Models. (b) 1) Previous methods predict annotation mask in only single step; 2) MedSAM-U, our method, automatically refine prompts with uncertainty-guided multi-prompt adaptation for reliable MedSAM. Example is from Colonoscopy test set~\cite{pogorelov2017kvasir}.}
\label{Figure_1}
\end{figure}

\section{Introduction}
\IEEEPARstart{M}{edical} image segmentation is an important task in medical image analysis, crucial for many clinical applications. Accurate segmentation helps clearly define anatomical structures and diseased areas, which is essential for disease diagnosis, treatment planning, and monitoring. It is widely used in dermoscopy, CT, MRI, colonoscopy, and ultrasound. Traditional U-Net-based segmentation methods have demonstrated high segmentation performance~\cite{Unet15,Unetadd20,isensee2021nnu,gao2021utnet}. However, these models are mostly designed for specific tasks and often face challenges in transferability to other domains.
Consequently, with the recent rise of foundation models like the Segment Anything Model (SAM)~\cite{kirillov2023segment}, and MedSAM~\cite{ma2024segment} have shown its universal applicability and precise segmentation capabilities across various datasets. Crucially, these models are trained on all datasets together and utilize different prompts to achieve more accurate segmentation results. As mentioned in~\cite{zhou2023edgesam}, high-quality prompts can lead to good performance. Hence, this study focuses on how to automatically obtain effective prompts during testing time without the need to retrain model.

As shown in Fig.~\ref{Figure_1} (a), we illustrate the segmentation performance of SAM and MedSAM across five datasets using various box prompts, including prompts with different aspect ratios, and positions within the image. The purpose of these variations in box prompts is to reveal a significant impact of box position on the resulting segmentation outcomes.  
Furthermore, as detailed in~\cite{zhou2023edgesam}, the effect of point prompts on segmentation performance is demonstrated. Therefore, the other key focus of this study is to train MedSAM to adapt to different types of prompts. Finally, as shown in Fig.~\ref{Figure_1} (b), MedSAM's approach of producing a single segmentation result in a single step lacks reliability. This study aims to leverage reliability to automatically obtain effective prompts, thereby achieving better segmentation results.

Uncertainty is one of crucial metrics for assessing the model's confidence or reliability. Existing methods for uncertainty estimation encompass dropout-based methods~\cite{17dropoutCV}, ensemble-based methods~\cite{ensemble17}, entropy-based methods~\cite{luo2019applicability}, evidential-based approaches~\cite{zou2023evidencecap}, and deterministic-based methods~\cite{ICMLdeterministic20}. Given the extensive parameters involved in MedSAM, building a new model with uncertainty estimation from scratch would demand significant computational resources and time. Furthermore, a key aspect of this study is exploring how to utilize estimated pixel-level uncertainty to obtain reliable prompts of different types.

To address these challenges, our study introduces the MedSAM-U as illustrated in Fig.~\ref{Figure_1} (b) 2), an automatic uncertainty-guided auto multi-prompt framework for adapting reliable MedSAM. MedSAM-U integrates box and point prompts to enhance segmentation accuracy. Subsequently, it then employs Uncertainty-Guided Multi-Prompt (UGMP) to effectively estimate the uncertainties associated with the prompts and their initial segmentation results. Our approach further introduces a novel Uncertainty-Guided Prompt Adaptation (UGPA) technique, enabling the auto refinement of multi-prompt to enhance segmentation reliability and accuracy. The contributions of this work are summarized as follows.\\ 
$\bullet$ We propose MedSAM-U, which leverages uncertainty estimation and guidance to automatically predict reliable segmentation results. To the best of our knowledge, this is the first attempt at using uncertainty-guided adaptation of multi-prompt to achieve a reliable MedSAM.\\
$\bullet$ We employ Uncertainty-Guided Multi-Prompt (UGMP) to estimate the uncertainty of different prompts in MedSAM without requiring additional training parameters.\\
$\bullet$ We introduce Uncertainty-Guided Prompt Adaptation (UGPA) to automatically obtain reliable prompts using the estimated uncertainty in the testing time, leading to accurate segmentation predictions.\\
$\bullet$ We conducted unified training and testing on five different modalities (Dermoscopy, Colonoscopy, Ultrasound, CT, and MRI) datasets, demonstrating the reliability and accuracy of MedSAM-U~\footnote{Our code will be released in https://github.com/Zhounan1222/MedSAM-U}.

\section{Related works}
\subsection{SAM for medical image segmentation}
Traditional deep learning methods~\cite{long2015fully,Unet15,chen2021transunet} are mostly designed for specific tasks and often face challenges in transferring to other domains. SAM~\cite{kirillov2023segment} is the pioneering large foundation model for segmentation, consists of three primary components: an image encoder, a prompt encoder, and a mask decoder. The image encoder is based on a standard Vision Transformer (ViT) that has been pretrained using Masked Autoencoders (MAE). The prompt encoder can operate in either a sparse (e.g.,boxes) or dense (e.g., masks) manner. The mask decoder is a Transformer decoder block adapted to include a dynamic mask prediction head. This decoder employs two-way cross-attention mechanisms to capture the interactions between the prompt and image embeddings. Subsequently, SAM upsamples the image embedding, subsequently, SAM upsamples the image embedding, and a Multi-Layer Perceptron (MLP) maps the resulting token to a dynamic linear classifier that predicts the ground truth (GT) for the given image $\textit{I}$. Due to its strong zero-shot performance, SAM marks a major breakthrough in the field of image segmentation. 

To improve the unsatisfactory performance of SAM on medical image segmentation tasks, some
approaches are to fine-tune SAM on medical images, including full fine-tuning and parameter-efficient fine-tuning~\cite{cheng2023sam,cui2024all, gong20243dsam,li2024prism}. Recently, MedSAM~\cite{ma2024segment} has investigated SAM’s application to medical image segmentation, exploring its performance in various contexts such as endoscopic surgery~\cite{cui2024surgical}, tumor
segmentation~\cite{hu2023sam}, polyp
segmentation~\cite{li2024polyp}. However, the existing MedSAM-based methods rarely investigate the sensitivity of different prompts.

\subsection{Prompts-based methods for medical image segmentation}
Prompts-based methods have gained traction in medical image segmentation due to their ability to provide flexible and adaptive guidance during interactive segmentation tasks. While effective in some cases, current adaptations of SAM rely heavily on high-quality, standard prompts (such as points, boxes, and masks) to deliver satisfactory performance in medical image segmentation tasks. Wu \textit{et al.}~\cite{wu2023medical} introduce Space-Depth Transpose for adapting 2D SAM to 3D medical images and Hyper-Prompting Adapter for prompt-conditioned adaptation. Deng \textit{et al.}~\cite{deng2023sam} propose a multi-box prompt-triggered uncertainty estimation for SAM that uses Monte Carlo methods and test-time augmentation to enhance performance and provide pixel-level reliability for segmented lesions or tissues. Li \textit{et al.}~\cite{li2023promise} propose a 3D medical image segmentation model using a single point prompt with SAM's pretrained vision transformer and lightweight adapters, featuring a hybrid network and boundary-aware loss for precise results. Wu \textit{et al.}~\cite{wu2024one} presents One-Prompt Segmentation method that combines one-shot and interactive approaches to handle unseen tasks with a single prompt in one pass. 
 Currently, most MedSAM-based methods are limited to using a single type of prompt rather than multiple types of prompts~\cite{yao2023false,zhang2023segment,deng2023sam,li2023promise}. This reduces the available information for the model and may result in insufficient segmentation performance.

\subsection{Uncertainty-based methods for medical image segmentation}
Uncertainty estimation~\cite{gal2016dropout} in segmentation has become increasingly important, as it offers a way to assess the confidence in model predictions. This is crucial in high-stakes applications such as medical image analysis, where mistakes can lead to serious repercussions. There are two primary types of uncertainty—aleatoric, which originates from the inherent noise in the data~\cite{hora1996aleatory}, and epistemic, which is due to the model's limitations—is the key to enhancing the reliability of deep neural networks~\cite{hullermeier2021aleatoric}. For instance, Saad \textit{et al.}~\cite{saad2010exploration} utilized shape and appearance priors to quantify uncertainty in probabilistic medical image segmentation. Parisot \textit{et al.}~\cite{parisot2014concurrent} leveraged segmentation uncertainty to inform adaptive sampling strategies for the simultaneous segmentation and registration of brain tumors. Additionally, Prassni \textit{et al.}~\cite{prassni2010uncertainty} developed a method to visualize uncertainty in random walker-based segmentation, which was then applied to volumetric segmentation of brain MRI and abdominal CT images. Zou \textit{et al.}~\cite{zou2022tbrats}. proposed a trusted brain tumor segmentation network that generates robust segmentation results and reliable uncertainty estimations by leveraging subjective logic theory to model uncertainty and parameterize class probabilities as a Dirichlet distribution.

Additionally, uncertainty estimation in large models, such as SAM, is critical for improving the reliability of segmentation outputs~\cite{deng2023sam}. 
Yao \textit{et al.}~\cite{yao2023false} propose a test-phase prompt augmentation technique for SAM that integrates multi-box augmentation with an aleatoric uncertainty-based FN and FP correction strategy to improve medical image segmentation. Zhang \textit{et al.}~\cite{zhang2023segment} propose UR-SAM, an uncertainty-rectified framework that enhances SAM's reliability in medical image segmentation by combining prompt augmentation with uncertainty-based rectification. Despite these advancements, research on integrating uncertainty estimation with prompt-based MedSAM remains insufficient, particularly in how to utilize uncertainty to guide reliable prompts.

\begin{figure*}[htbp]
\centering
\includegraphics[width=1\linewidth]{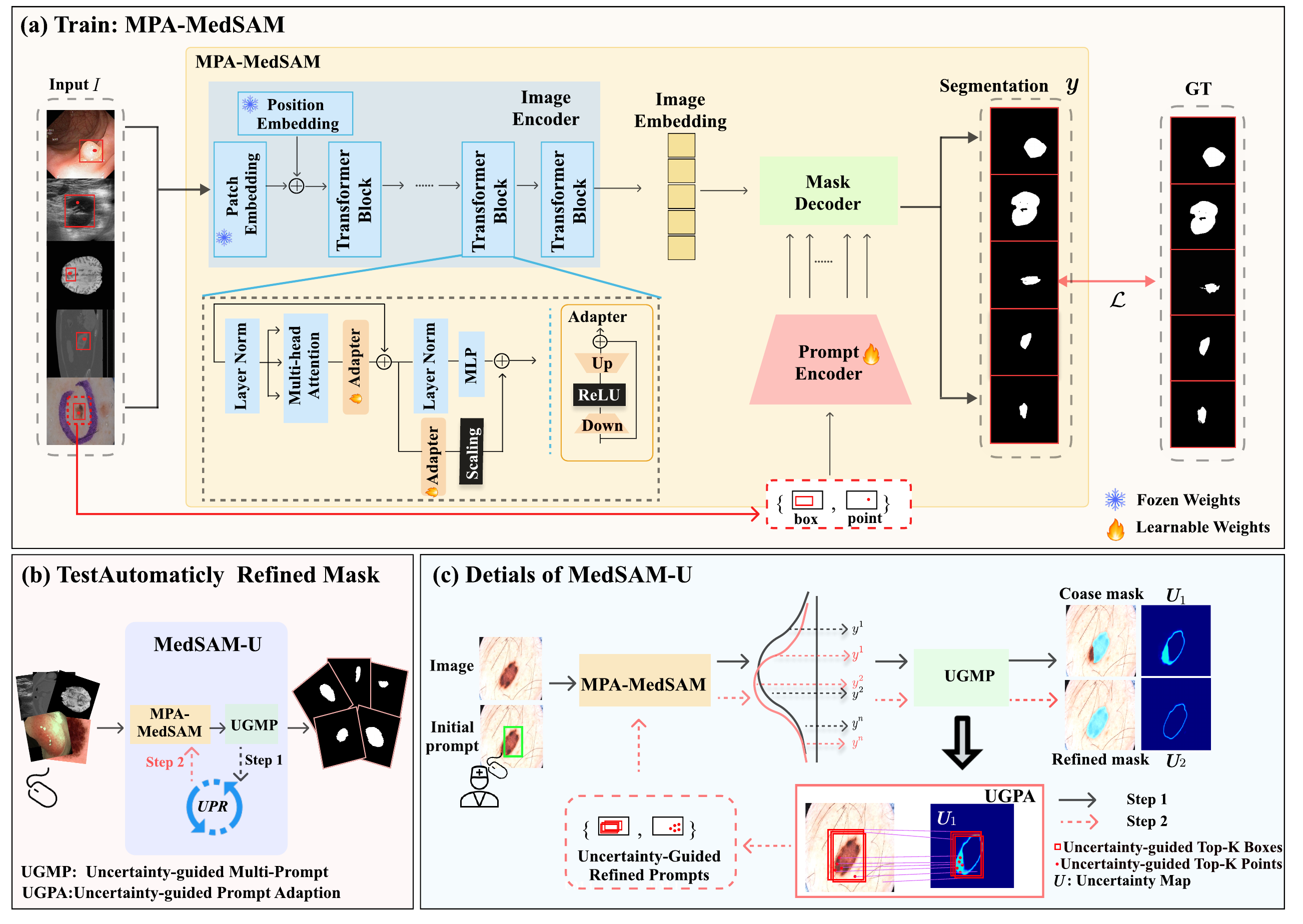}
\centering \caption {The overall framework of MedSAM-U. The framework is presented through three key illustrations: (a) the training process of MPA-MedSAM, (b) a comprehensive work flow in the inference time, and (c) a simplified diagram that illustrates the user interaction process.}
\label{F_2}
\end{figure*}

\section{Method}
To begin with, we provide an overview of the MedSAM-U architecture, which consists of three primary components: MPA-MedSAM, UGMP, and UGPA modules. The overall framework of our proposed method is illustrated in Fig.~\ref{F_2}. We build an automatic framework guided by uncertainty, developed to adaptively refine multi-prompt to achieve reliable and accurate segmentation that includes 1) training Multi-Prompt Adaption (MPA) to integrating both point and box cues to achieve more precise medical image segmentation results. 2) In the reference time, utilizing UGMP to assess the uncertainty linked to various prompts, without requiring additional training parameters. 3) introducing UGPA to automatically leverage estimated uncertainty to derive more reliable prompts, resulting in improving the peformance of segmentation results.

\subsection{Multi-Prompt Adaptation for MedSAM}
MedSAM~\cite{ma2024segment} primarily relies on box prompts as the initial input for segmentation tasks, without incorporating point prompts during its training process. This limitation means that MedSAM's segmentation capabilities are optimized for scenarios where box annotations are available, but it may not fully exploit the potential benefits of point-based inputs. Formally, in MedSAM, the relationship between the inputs and the prediction mask $\boldsymbol{y}$ can be formally expressed as:
\begin{equation}
{\boldsymbol y} = f_{\text{MSAM}}(\textit{I}, b),
\label{E_1}
\end{equation}
the function \( f_{\text{MSAM}} \) can be denoted as:
\begin{equation}
f_{\text{MSAM}}(\textit{I}, b)=\mathcal{F}_D\left(\mathcal{F}_E(\textit{I}), \mathcal{F_P}(b)\right),
\label{E_2}
\end{equation}
where \( \textit{I} \) denotes the input image, \( b \) represents the box prompt, \( \mathcal{F}_D\), \( \mathcal{F}_E \) and \( \mathcal{F_P} \) represent the Decoder, Image Encoder, Prompt Encoder modules of MedSAM, respectively, and \( \boldsymbol{y} \) is the resulting segmentation mask of MedSAM. 

Moreover, as detailed in~\cite{zhou2023edgesam,li2024prism}, the effect of point prompts on segmentation performance is demonstrated, particularly positive points. Accordingly, to improve the sparse encoder, this study proposes modifications that refine the encoding of points and boxes. These adjustments aim to optimize the integration of positional encoding with learned embeddings for each prompt type. We propose MPA that is designed to handle and adapt multiple types of prompt inputs for MedSAM, including the combination of points and boxes prompts. 

Our goal is to extend the capabilities of MedSAM by adapting it to handle multi-prompt through a fine-tuning approach, called MPA-MedSAM, designed to enhance MedSAM's flexibility without the need for full re-training. Instead of adjusting all parameters, we keep the pre-trained MedSAM parameters frozen except for the prompt encoder, develop an adapter module, and integrate it into the image encoder designated positions. Structurally, the adapter serves as a bottleneck model, consisting of three key components: a down-projection, ReLU activation, and up-projection sequentially, as illustrated in Fig.~\ref{F_2} (a), similar to Med-SA~\cite{wu2023medical}. Given that MedSAM does not show improvements when multi-type prompts are incorporated, we unfreeze the prompt encoder to expand its capabilities. To break the limitation, we have unfrozen the prompt encoder to expand its capabilities. This adjustment allows MedSAM to effectively handle multi-type prompts, addressing the limitations encountered in earlier versions. For interactive segmentation, both point prompts and Bounding box (BBox) prompts are utilized during training. The prompts are generated by randomly selecting points and applying jitter to the Bbox derived from the segmentation mask. This simulates varying levels of user inaccuracy, making the model more robust to real-world variations and enhancing its adaptability in practical applications. The relationship between the inputs and the prediction mask $\boldsymbol{y}$ in MPA-MedSAM can be formally expressed as:
\begin{equation}
{\boldsymbol{y}} = f_{\text{MPA-MSAM}}(\textit{I}, p, b),
\label{medsam-adapter}
\end{equation}
where \( \textit{I} \) denotes the input image, \( p \) represents the point prompt, \( b \) represents the Bbox prompt, and \( \boldsymbol{y} \) is the resulting segmentation mask of MPA-MedSAM. The function \( f_{\text{MPA-MedSAM}} \) encapsulates the MPA-MedSAM, which processes the input image, point prompt, and box prompt to produce the corresponding mask.

\subsection{Uncertainty estimation of MPA-MedSAM with multi-type prompts}
Given the variability of prompts from users with differing levels of experience, MedSAM's segmentation performance exhibits significant sensitivity to the position of Bboxes, potentially leading to inference errors. To address this, we adopt a strategy similar to the one used by SAM-U~\cite{deng2023sam}. In this study, we introduce UGMP to incorporate multiple prompts to improve the accuracy of MPA-MedSAM. 
Assume giving an initial Bbox \(b^{\text{init}}\), by applying a simple random augmentation strategy to the \(b^{\text{init}}\), we generate $N$ multiple Bbox prompts as follows: 
\begin{equation}
\{b_1, b_2, \dots, b_N\} = \{b^{\text{init}} + \delta_1, b^{\text{init}} + \delta_2, \dots, b^{\text{init}} + \delta_N\}, 
\label{permute}
\end{equation}
where $\delta_i \sim \mathcal{N}(\mu, \sigma)$, $b^{\text{init}}$ represents the initial Bbox, these random boxes are generated by applying adjustments \(\delta_i\) to \(b^{\text{init}}\), $\mathcal{N}$ denotes the normal distribution, $\mu$ denotes the coordinates of the $b^{\text{init}}$, $\delta$ represent the degree of perturbation applied to the $b^{\text{init}}$. The operation generate a set of multiple Bbox prompts $\mathbb{B} = \{b_1, b_2, \dots, b_N\}$ , and a set of $M$ multiple point prompts $\mathbb{P} =\{p_1, p_2, \dots, p_M \}$ that depends on the user's or clinician's choice, $\mathbb{P}= \emptyset$ if no points are used. To quantify the uncertainty arising from the use of multiple prompts, with $N$ box prompts, $M$ point prompts and input image $\textit{I}$, MPA-MedSAM predict a set of results $ \mathbb{Y}=\{{\boldsymbol{y}}_1,{\boldsymbol{y}}_2,\cdots,{\boldsymbol{y}_N}\}$, where $\boldsymbol{y}_i$ is predicted as follows:
\begin{equation}
\boldsymbol{y_i} = {{ f_{\text{MPA-MedSAM}}}\left( {\textit{I}},{\mathbb{B}_i}, {\mathbb{P}} \right)}, 
\label{u-medsam-adapter_i}
\end{equation}
where $b_i$ represents the \(i\)-th box of the set $\mathbb{B}$, $y_i$ represents the segmentation result obtained by inputting $b_i$ into MedSAM.

The aggregation of these predictions leads to enhanced segmentation accuracy and a reduction in uncertainty. The final combined prediction is formulated as follows:
\begin{equation}
\overline{\boldsymbol{y}} = \frac{1}{N}\sum\limits_{i = 1}^N {\boldsymbol{y_i}},  
\label{u-medsam-adapter}
\end{equation}
\( \overline{\boldsymbol{y}} \) represents the average segmentation result obtained by applying the MPA-MedSAM model across \( N \) instances.

By ultilizing UGMP, the aleatoric uncertainty from a single given image $I$ , instead of being estimated from each individual prediction \( \boldsymbol{y}_i \), is now directly estimated from the average prediction \( \overline{\boldsymbol{y}} \),  described by the entropy~\cite{bein2006entropy} :
\begin{equation}
\overline{\textbf{\textit{U}}} = f_{\text{UGMP}}\left( \overline{\boldsymbol{y}} \right),
\label{E_4}
\end{equation}
where $f_{\text{UGMP}}\left( \overline{\boldsymbol{y}} \right) = -\int p(\overline{\boldsymbol{y}}|\textit{I}) \log{p(\overline{\boldsymbol{y}}|\textit{I})} d\overline{\boldsymbol{y}}$. This allows us to estimate the uncertainty distribution based on the aggregated prediction \( \overline{\boldsymbol{y}} \), rather than calculating it for each individual prediction \( \boldsymbol{y}_i \).

In summary, the MPA-MedSAM with UGPM function can be expressed as follows: 
\begin{equation}
(\overline{\boldsymbol{y}}, \overline{\textbf{\textit{U}}}) = f_{\text{UGMP}}(f_{\text{MPA-MedSAM}}\left(\textit{I}, \mathbb{B}, \mathbb{P} \right)),
\end{equation}
where \(\textit{I} \) represents the input image, $P_b$ are the multiple Bbox prompts \( \{b_1, b_2, \dots, b_N\} \), $P_p$ are the points prompts \( \{p_1, p_2, \dots, p_M\} \), also depending on the user's or clinician's choice. \(\overline{\boldsymbol{y}} \) is the resulting segmentation mask based on the combination of predictions from multiple prompts, and \(\overline{\textbf{\textit{U}}} \) is the associated uncertainty map.

\subsection{Uncertainty-Guided Prompts Adaptation}
Providing prompts that produce desired results can be a difficult process that often requires the users or experts to go through tedious trial-and-error experimentation. Due to the uncertainty introduced by different prompts during the inference process, we investigated whether this knowledge could be utilized to refine the initial prompts, thereby producing more predictable and precise segmentation mask outputs. In an effort to address this issue, We introduce UGPA over a series of sample prompts to refine them, through UGPA could improve the model’s segmentation through automatic processes. 
Fig.~\ref{F_2} (b) shows the schematic view of our proposed UGPA. 

In the UGPA process, firstly, we select the Top-\textit{K} Bboxes based on the edges of the $\textbf{\textit{U}}$, which highlight regions of high uncertainty. These boxes are then subjected to slight adjustments, simulating expert adjustments to refine their accuracy. Secondly, following this, we select the Top-\textit{K} points with the highest uncertainty values from the $\textbf{\textit{U}}$, representing the most uncertain areas within the image. and then these selected points are then combined with the adjusted Bboxes to create refined prompts. To provide a clear and intuitive description of the process, we represent each step using the following formulas:
\begin{equation}
b_{i} = \left [(w_{\text{min}}+ \delta_i, h_{\text{min}}+ \delta_i), (w_{\text{max}}+ \delta_i, h_{\text{max}}+ \delta_i) \right],
\label{Eb}
\end{equation}
where:
\begin{equation}
(w_{\min}, h_{\min}) = \min_{(w_j, h_j) \in \text{Edge}(\textbf{\textit{U}})} (w_j, h_j),
\end{equation}
\begin{equation}
(w_{\max}, h_{\max}) = \max_{(w_j, h_j) \in \text{Edge}(\textbf{\textit{U}})} (w_j, h_j),
\end{equation}
$\text{Edge}(\textbf{\textit{U}})$ refers to identifying the coordinates of the edges of the uncertainty map, \((w_j, h_j)\) refers to the coordinates of the edge points. Specifically, \((w_{\min}, h_{\min})\) represents the coordinates of the top-left corner, and \((w_{\max}, h_{\max})\) represents the coordinates of the bottom-right corner of the bounding box that encloses these edge points. We select the Top-1 Bbox based on the edges of the $\textbf{\textit{U}}$, which has the largest area and fully contains the edge of $\textbf{\textit{U}}$. Then, we also apply Eq. (4) to simulate varying levels of expert knowledge, to generate a set of refined box prompts $\mathbb{B}^* = \{b_1, b_2, \dots, b_k\}$ . Then, we select the Top-\textit{K} points with the highest uncertainty values from the $\textbf{\textit{U}}$:
\begin{equation}
\{p_1, p_2, \dots, p_k\} = \mathcal{S}\left(\textbf{\textit{U}}\right)_{[\,:k]},
\label{E_p}
\end{equation}
$\mathcal{S}$ denotes the sort function, $\{p_1, p_2, \dots, p_k\}$ represents a set of the selected $k$ points $\mathbb{P}^*$.
we combine the adjusted Bboxes with the selected points to create the refined prompts set: $\{ \mathbb{B}^*, \mathbb{P}^* \}$.

Subsequently, these refined prompts are re-as-input into MPA-MedSAM, where they serve as crucial inputs to guide further refinement of the segmentation output. By leveraging these improved prompts, our method can enhance the accuracy of the segmentation, narrowing down errors and producing more precise results. This process ensures that the model benefits from the enhanced guidance provided by the refined prompts, ultimately leading to superior segmentation performance. The refined prompts then input into MPA-MedSAM with UGPM function can be expressed as follows:
\begin{equation}
({\boldsymbol{y}}^*, {\textbf{\textit{U}}}^*) = f_{\text{UGMP}}(f_{\text{MPA-MedSAM}}\left(\textit{I}, \mathbb{B}^*, \mathbb{P}^* \right)),
\end{equation}
$\boldsymbol{y}^*$ is the segmentation mask based on the UPGA, and $\textbf{\textit{U}}^*$ is the associated uncertainty map.

Finally, the refined process is informed by Active Learning principles: we assess the output uncertainty following the prompt optimizations and update the segmentation results only if these optimizations lead to reduced uncertainty estimates. Importantly, the final output is not considered definitive until it is rigorously verified against the uncertainty map to ensure that performance has indeed improved. It can be denoted by:
\begin{equation}
    \hat{\boldsymbol{y}} = \begin{cases} 
    {\boldsymbol{y}}^* & \text{if } \overline{\textbf{\textit{U}}} > {\textbf{\textit{U}}}^*, \\
    \overline{\boldsymbol{y}} & \text{otherwise}.
    \end{cases}
\end{equation}
The detials of the proposed method MedSAM-U are outlined in Algorithm.~\ref{alg:UGPA}.
\begin{algorithm}
\caption{The proposed method MedSAM-U}
\label{alg:UGPA}
    \textbf{Input:} Image $ \textit{I} $, Initial Bbox $b^{\text{init}}$
    
    \textbf{Output:} Refined mask  $\hat{\boldsymbol{y}}$
    
    1 Generate random boxes based on $b^{\text{init}}$ with Eq. (4):
    \[
    \mathbb{B} \xleftarrow{} b^{\text{init}}
    \] 
    
    2 Input to MedSAM-U with Eq. (8):
    \[
     (\overline{\boldsymbol{y}}, \overline{\textbf{\textit{U}}}) \xleftarrow{} \mathbb{B}
    \] 
    
    3 Select Top-\textit{K} Boxes and Points with Eq. (9) to Eq. (12):
    \[
    \{\mathbb{B}^*,\mathbb{P}^*\} \xleftarrow{} \overline{\textbf{\textit{U}}}
    \]

    4  Re-input to MedSAM-U with Eq. (13):
    \[
    ({\boldsymbol{y}}^*, {\textbf{\textit{U}}}^*)  \xleftarrow{} \{\mathbb{B}^*,\mathbb{P}^*\}
    \]

    5 Verify and Update with Eq. (14)

    6 \textbf{Output:} $\hat{\boldsymbol{y}}$
\end{algorithm}
    
\section{Experiments and Results}
\subsection{Datasets \& Loss \& Implementation Details}
\subsubsection {Datasets} 
To assess the effectiveness of our proposed method MedSAM-U, we choose five different 2D medical imaging modalities, including Dermoscopy, CT, MRI, Colonoscopy, Ultrasound. Each modality is represented by a single dataset, with the specific details provided below.

\noindent
\textbf{Dermoscopy}: ISIC-2017~\cite{gutman2016skin} , hosted at the Medical Image Computing and Computer Assisted Intervention (MICCAI) conference, is a skin lesion segmentation dataset towards melanoma detection, including 2594 annotated images.

\noindent
\textbf{Ultrasound}: CT2US~\cite{song2022ct2us} is a dataset designed for kidney segmentation using cross-modal transfer learning. It includes paired CT and ultrasound images with corresponding annotations for kidney segmentation, aiming to improve segmentation accuracy in ultrasound images with limited data.The dataset has a total of 4,586 samples and is simply called Kidney.

 \noindent
\textbf{CT}: This research study used open access segmented data KiTS23~\cite{heller2023kits21}. Since the dataset is primarily designed for 3D segmentation tasks, we adapted it to perform 2D segmentation. To achieve this, we converted the 3D volumetric data into 2D slices. Specifically, we extracted slices along the z-axis, focusing on the central region of each volume, and selected slices at regular intervals to ensure a representative and manageable dataset which has a total of 3882 samples.

\noindent
\textbf{MRI}: In this study, we utilize the publicly available BraTS 2021 glioma segmentation MRI dataset for evaluation purposes. Since only the training dataset includes GT segmentation masks, making it suitable for an automatic evaluation of a point-to-mask task similar to SAM~\cite{kirillov2023segment}, we exclusively use the training dataset for our current evaluation. To assess our method's potential in supporting interactive clinical treatment planning, and due to the model's limitation to a single image input, we evaluated segmentation accuracy using the T1 modality MRI sequence as input. We extracted all slice images and their corresponding masks along the z-axis, selecting middle slices at intervals to convert the 3D images into 2D slices to generate a total of 4,586 samples Subsequently, we randomly split the data, using 0.8 for training and 0.2 for testing, based on image indices.

\noindent
\textbf{Colonoscopy}: we use the Kvasir-SEG dataset~\cite{pogorelov2017kvasir}, which consists
of 1,000 polyp images and their corresponding GT masks annotated by expert endoscopists from Oslo University Hospital (Norway).

All the data are used for evaluation. The $N$ is set to 3 for our experiments. Box prompts were generated based on the area and size of the ground truth. The length and width of the boxes were randomly adjusted to mimic the manually provided box prompts.

\subsubsection {Loss function} 
In our method, for training the MPA to enable MedSAM to adapt to various types of prompts, we utilize a combination of a binary focal loss function~\cite{lin2017focal} and dice loss functions~\cite{jadon2020survey} to supervise the output effectively during the training process, following the~\cite{kaur2023improving}. This combined loss function is designed to address the challenges of class imbalance and accurate boundary delineation in segmentation tasks. The loss is calculated by the following formula:

\begin{equation}
    \mathcal{L}= \left[ -\alpha \sum_{t}^N (1 - y_t)^\gamma \log(y_t) \right]+\left[ 1 - \frac{2 \sum_{t}^N y_t \cdot g_t}{\sum_{t}^N y_t + \sum_{t}^N g_t} \right] ,
\end{equation}
where:
\begin{equation}
    y = f_{\text{MPA-MedSAM}}(\textit{I}, b,p; \theta),
\end{equation}
\(\textit{I}\) denotes the input image, $f_{\text{MPA-MedSAM}}$ represents the MPA-MedSAM predict the segmentation result $y$,  with the Bbox prompt $b$ and point prompt $p$. $y_t$ denotes its probability for pixel, and \(\theta\) represents the model parameters needed to update, \(g_t\) denotes the ground truth label for pixel \(t\).

\subsubsection {Implementation Details}

For interactive segmentation, we employ Bbox prompts combined with points prompts during the model training process. In this study, we adhered to the default training settings of MedSAM for 2D medical image training. Taking five different modality datasets as input, we trained the model for 60 epochs. We chose a smaller number of epochs compared to fully fine-tuned training. In the interactive model, during the initial step of simulating user clicks to initialize prompts, we experimented with various prompt settings. These included: (1) a random 3 positive points, denoted as \textbf{3P}, (2) a random 5 positive points, denoted as \textbf{5P}, (3) a random 10 positive points, denoted as \textbf{10P}, (4) 3 Bboxes with 50\% overlapping of the target, denoted as \textbf{3B (0.5)}, and (5) 3 Bboxes with 75\% overlapping of the target, denoted as \textbf{3B (0.75)}, (6) the multi-type prompts composed of both 3 points and 3 Bboxes annotations of 0.5 or 0.75 are referred to as the \textbf{3P \& 3B (0.5)}, and \textbf{3P \& 3B (0.75)}. Similarly, the same approach can be applied to other cases.
All the experiments are implemented with the PyTorch platform and trained/tested on a single NVIDIA 4090 GPU. We utilized the default settings to reproduce the comparison methods. We use three commonly-used metrics for the evaluation: Dice Coefficient (Dice) and Intersection over Union (IoU). Dice calculates the overlap between the prediction and GT as twice the area of overlap divided by the sum of the areas of the prediction and GT. IoU measures the ratio of the intersection to the union of the predicted and true regions. 
\begin{table*}[htbp]
    \centering
    \setlength{\tabcolsep}{8pt}

    \caption{ Results comparing our method with other segmentation methods across five datasets are evaluated using Dice Score and IOU Score. Here, SAM 3B (0.5) refers to using 3 low-quality initial Bboxes with the overlap ratio of 0.5 as inputs to the SAM model. The same approach applies to other cases. The top-2 results are highlighted in \textbf{bold} and \underline{underline}.}
    \label{tab:baseline}
    \begin{tabular}{lccccccccccccc} 
        \toprule
        \multirow{2}{*}{ \textbf{Model}}  & \multicolumn{2}{c}{\textbf{Dermoscopy}}  & \multicolumn{2}{c}{\textbf{Colonoscopy}} & \multicolumn{2}{c}{\textbf{Ultrasound}} & \multicolumn{2}{c}{\textbf{CT}} & \multicolumn{2}{c}{\textbf{MRI}} & \multicolumn{2}{c}{\textbf{Avg}} \\
        \cmidrule(lr){2-3} \cmidrule(lr){4-5} \cmidrule(lr){6-7} \cmidrule(lr){8-9} \cmidrule(lr){10-11} \cmidrule(lr){12-13}
        & {IoU} & {Dice} & { IoU} & { Dice} & { IoU} & { Dice} & { IoU} & { Dice} & { IoU} & { Dice}
        & { IoU} & { Dice} \\
        
        \midrule
        \textbf{SAM 3B (0.5)} & 0.481	& 0.609 & 0.346 & 0.414 & 0.467	& 0.619 & 0.558	& 0.665 & 0.233	& 0.321 & 0.417	& 0.526 \\
        \textbf{SAM 3B (0.75)}  & 0.656	& 0.773	& 0.651	& 0.725	& 0.656	& 0.783	& 0.667	& 0.762	& 0.470	& 0.596	& 0.620	& 0.728  \\
        \textbf{MedSAM 3B (0.5)} & 0.446	& 0.566	& 0.516	& 0.646	& 0.566	& 0.705	& 0.455	& 0.585	& 0.516	& 0.665	& 0.500	& 0.633 \\
        \textbf{MedSAM 3B (0.75)}  & 0.778	& 0.861	& \underline{0.811}	& 0.880	& 0.873	& 0.931	& 0.661	& 0.763	& \textbf{0.664}	&  \textbf{0.785}	& \underline{0.758}	& \underline{0.844} \\
        \midrule
        \textbf{MedSAM-U 3B (0.5)} & \underline{0.801}	& \underline{0.883}	& 0.779	& 0.867	& \underline{0.899}	& \underline{0.946}	& \underline{0.672}	& \underline{0.766}	& 0.599	& 0.727	& 0.750	& 0.838 \\
        \textbf{MedSAM-U 3B (0.75)} & \textbf{0.839}	& \textbf{0.909}	& \textbf{0.833}	& \textbf{0.903}	& \textbf{0.920}	& \textbf{0.958}	& \textbf{0.700}	& \textbf{0.783}	& \underline{0.626}	& \underline{0.750}	& \textbf{0.784}	& \textbf{0.861} \\

        \bottomrule
    \end{tabular}
\end{table*}

\begin{figure*}[htbp]
    \centering
    \includegraphics[width=0.95\linewidth]{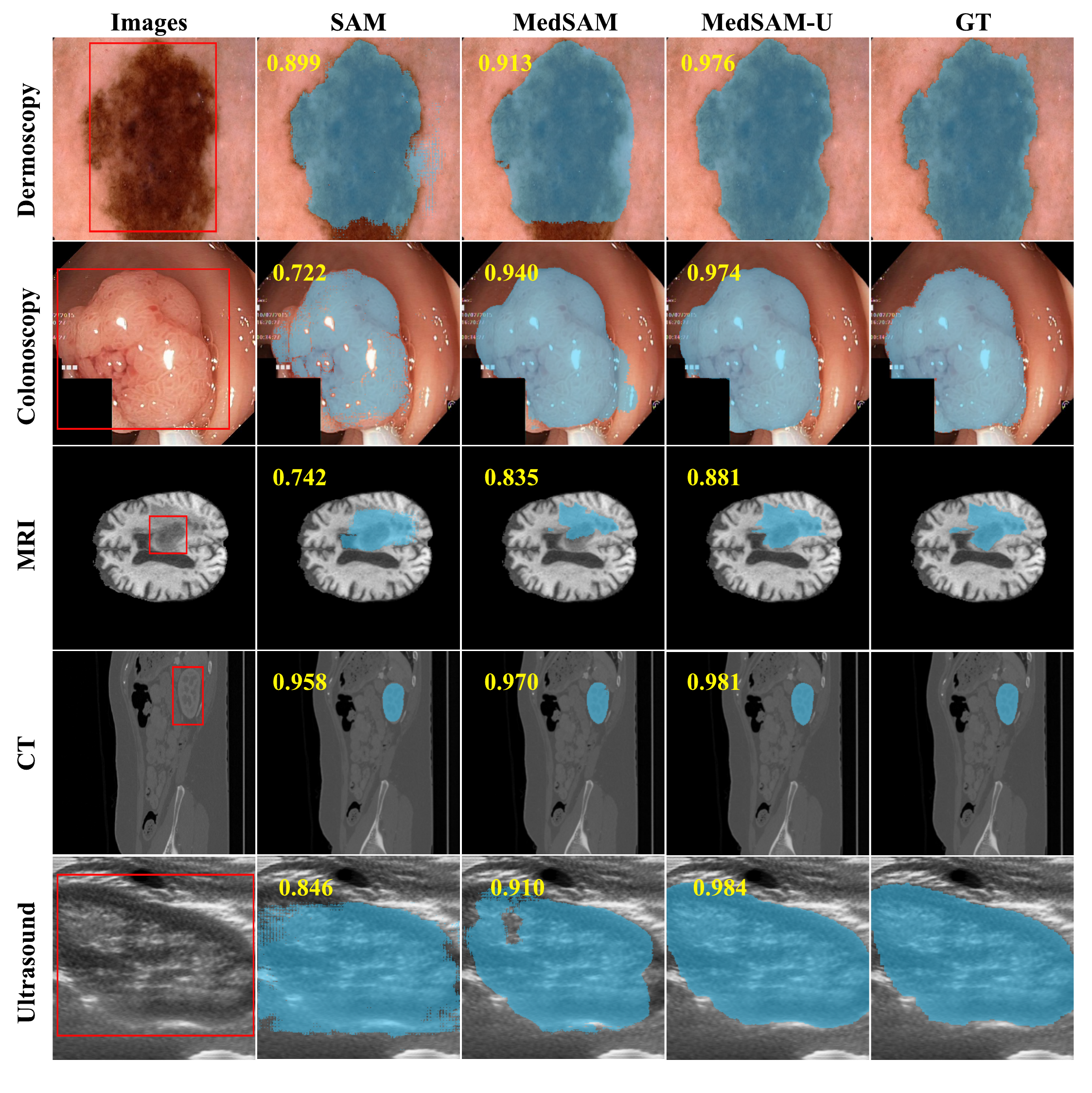}
    
    \caption {Visualization of segmentation results for each method in different modalities, with Bbox illustrating varying degrees of rough approximations simulating expert annotations. (Col. 1) Images
with an initial low-quality Bbox prompt; (Col. 2) SAM model; (Col. 3) MedSAM model; (Col. 4) Our model; (Col. 5) binary GT mask. Red: initial BBox, Blue: segmentation results Yellow: Dice Score. }
    \label{F_3}
\end{figure*}

\begin{figure*}[t]
    \centering
    \includegraphics[width=1\linewidth]{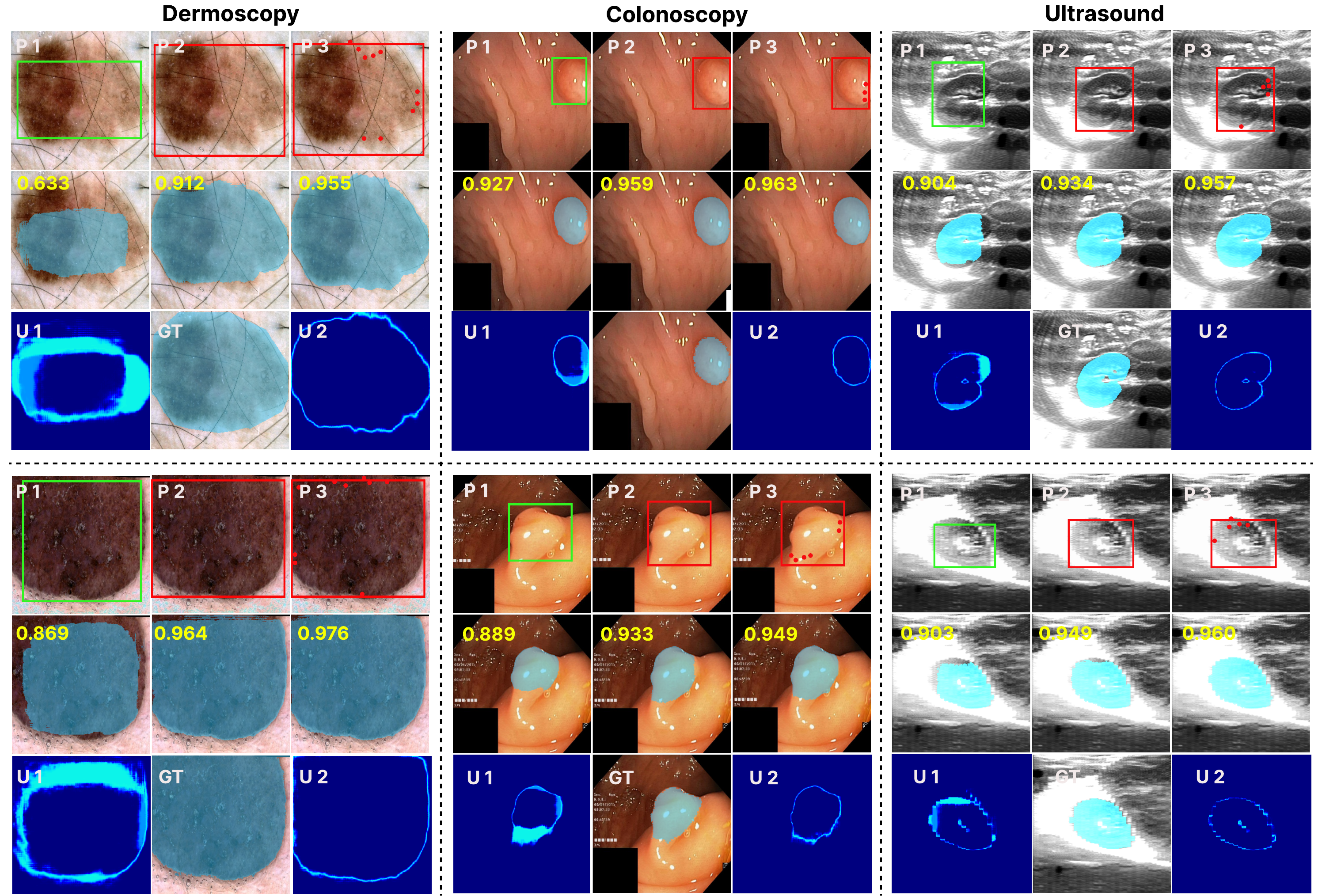}
    
    
    \caption {Visualization of segmentation results for our method in different modalities. P 1 : low-quality box prompts, P 2 : refined box prompts, P 3 : refined box and point prompts, U 1 : step 1 Uncertainty Map, U 2 : step 2 Uncertainty Map, GT : ground truth. Green: initial Bbox, Red: refined prompts, Blue: segmentation results, Yellow: Dice Score}
    \label{fig:F_4}
\end{figure*}

\begin{table*}[htbp]
    \setlength{\tabcolsep}{8pt}

    \centering
    \caption{ An comparative study on the performance of interactive segmentation models using different multi-type prompts. The results, evaluated using Dice Score and IoU Score across five datasets, are compared. Here, SAM 3P denotes using 3 points as inputs to the SAM model; SAM 3B (0.5) denotes using 3 low-quality initial Bboxes with the overlap ratio of 0.5 as inputs to the SAM model. The same approach applies to other cases. The top-1 results are highlighted in \textbf{bold}.}
    \label{tab:multi-prompt}
    \begin{tabular}{lcccccccccccccc} 
        \toprule
        \multirow{2}{*}{ \textbf{\fontsize{8}{9}\selectfont Model}} & \multicolumn{2}{c}{\textbf{\fontsize{9}{9}\selectfont Dermoscopy}} & \multicolumn{2}{c}{\textbf{\fontsize{9}{9}\selectfont Colonoscopy}} & \multicolumn{2}{c}{\textbf{\fontsize{9}{9}\selectfont Ultrasound}} & \multicolumn{2}{c}{\textbf{\fontsize{9}{9}\selectfont CT}} & \multicolumn{2}{c}{\textbf{\fontsize{9}{9}\selectfont MRI}} & \multicolumn{2}{c}{\textbf{Avg}} \\ 
        \cmidrule(lr){2-3} \cmidrule(lr){4-5} \cmidrule(lr){6-7} \cmidrule(lr){8-9} \cmidrule(lr){10-11} \cmidrule(lr){12-13}
        & {\fontsize{8}{8}\selectfont IoU} & {\fontsize{8}{8}\selectfont Dice} & {\fontsize{8}{8}\selectfont IoU} & {\fontsize{8}{8}\selectfont Dice} & {\fontsize{8}{8}\selectfont IoU} & {\fontsize{8}{8}\selectfont Dice} & {\fontsize{8}{8}\selectfont IoU} & {\fontsize{8}{8}\selectfont Dice} & {\fontsize{8}{8}\selectfont IoU} & {\fontsize{8}{8}\selectfont Dice}
        & {\fontsize{8}{8}\selectfont IoU} & {\fontsize{8}{8}\selectfont Dice} \\
        \midrule
         \textbf{SAM 3P}  & 0.375	& 0.490	& 0.283	& 0.359	& 0.229	& 0.360	& 0.074	& 0.112	& 0.127	& 0.209	& 0.217	& 0.306 \\
         \textbf{SAM 10P}  & 0.384	& 0.504	& 0.288	& 0.373	& 0.238	& 0.374	& 0.073	& 0.115	& 0.131	& 0.215	& 0.223	& 0.316 \\
        \textbf{SAM 3B (0.5)}  &  {0.481}	&  {0.609}	&  {0.346}	&  {0.414}	&  {0.467}	&  {0.619}	&  {0.558}	&  {0.665}	&  {0.233}	&  {0.321}	&  {0.417}	&  {0.526} \\
         \textbf{SAM 10P\&3B (0.5)}   & {0.504}	& {0.644}	& {0.368}	& {0.467}	& 0.344	& 0.504	& 0.182	& 0.280	& 0.229	& {0.347}	& 0.325	& 0.448 \\
         \textbf{MedSAM 3P}  & 0.060	& 0.106	& 0.036	& 0.063	& 0.093	& 0.159	& 0.016	& 0.029	& 0.027	& 0.047	& 0.046	& 0.081 \\
         \textbf{MedSAM 10P} & 0.146	& 0.240	& 0.078	& 0.130	& 0.149	& 0.245	& 0.028	& 0.048	& 0.060	& 0.100	& 0.092	& 0.153 \\
        \textbf{MedSAM 3B (0.5)} &  {0.446}	&  {0.566}	&  {0.516}	&  {0.646}	&  {0.566}	&  {0.705}	&  {0.455}	&  {0.585}	&  {0.516}	&  {0.665}	&  {0.500}	&  {0.633} \\
         \textbf{MedSAM 10P\&3B (0.5)}  & 0.341	& 0.467	& 0.281	& 0.388	& 0.153	& 0.240	& 0.270	& 0.386	& 0.284	& 0.406	& 0.266	& 0.377  \\
        \midrule
         \textbf{MPA-MedSAM 3P}  & 0.125	& 0.188	& 0.063	& 0.096	& 0.194	& 0.280	& 0.002	& 0.003	& 0.009	& 0.015	& 0.079	& 0.117  \\
         \textbf{MPA-MedSAM 10P}  & 0.266	& 0.374	& 0.113	& 0.174	& 0.371	& 0.497	& 0.005	& 0.009	& 0.014	& 0.024	& 0.154	& 0.216 \\
         \textbf{MPA-MedSAM 3B (0.5)}  & {0.623}	& {0.753}	& {0.554}	& {0.693}	& {0.678}	& {0.799}	& {0.479}	& {0.607}	& {0.400}	& {0.542}	& {0.547}	& {0.679} \\
         \textbf{MPA-MedSAM 10P\&3B (0.5)}  & \textbf{0.662}	& \textbf{0.784}	& \textbf{0.596}	& \textbf{0.729}	& \textbf{0.718}	& \textbf{0.829}	& \textbf{0.519}	& \textbf{0.644}& \textbf{0.441}	& \textbf{0.584}& \textbf{0.587}	& \textbf{0.714}  \\

        \bottomrule
    \end{tabular}
\end{table*}

\subsection{Comparisons with the SOTA methods on Multi-modality Images}
To validate the overall performance of our proposed method, we compared it with the SOTA segmentation foundation model across five different modalities. As shown in Table~\ref{tab:baseline}, include comparisons with SAM~\cite{kirillov2023segment}, and fully fine-tuned SAM in medical (MedSAM)~\cite{ma2024segment}, evaluated using Dice Score and Iou Score. The results were shown in Table~\ref{tab:baseline}.

Firstly, when comparing the performance of the same model under different Bboxes qualities (ratio from 0.5 to 0.75), a notable improvement is observed in SAM across all imaging modalities as the Bboxes ratio increases from 0.5 to 0.75. MedSAM exhibits a similar trend, with performance enhancements corresponding to the higher Bboxes quality. It reveals that the quality of the Bboxes significantly impacts segmentation performance, indicating that the models are sensitive to the quality of the Box prompts. 

Furthermore, for our proposed method, represented as MedSAM-U 3B (0.5), and 3 B (0.75), shown in the $5^{th}$ row and the $6^{th}$ row, achieved SOTA performance in Dermoscopy, Colonoscopy, Ultrasound and CT, with a final Avg Dice of 90.9 \% , 90.3 \% , 95.8\% and 78.3\% with 3B (0.75), surpassing MedSAM by 4.8\%, 2.3\%, 2.7\% and 2\% respectively. When the Bboxes overlap ratio is 0.5, our method outperforms surpassing MedSAM by 20.5 \%. The results shown in Table~\ref{tab:baseline} demonstrate that, even when the initial Bboxes quality is low, significant improvements in segmentation performance are observed. This demonstrates the effectiveness of our approach in enhancing segmentation accuracy, particularly in scenarios where the initial Bx prompts are suboptimal. These results reveal that our MedSAM-U demonstrates significant accuracy across various medical segmentation tasks with the use of low-quality Bboxes, eliminating the need for manual automantic adjustments to achieve satisfactory results. 

Fig.~\ref{F_3} visualized Some examples' segmentation results for each method in different modalities, with Bboxes simulating varying degrees of rough approximations simulating expert annotations. SAM and MedSAM's segmentations are based on a single step, and when the initial Bbox prompts provided by experts are not perfectly accurate, the segmentations may suffer from under-segmentation or over-segmentation errors. In contrast, MedSAM-U can accurately segment various targets across different imaging conditions.

\subsection{Ablation Study}

\subsubsection{Effectiveness of Multi-Prompt Combinations}

In our method,the MPA was introduced to integrate diverse types of prompts (e.g., points and boxes) into MedSAM, enabling the simultaneous input of various prompt combinations. To validate the effectiveness of the MPA-MedSAM module, we present the segmentation results of different interactive segmentation models using various prompt combinations across five different datasets, as detailed in Table~\ref{tab:multi-prompt}. 
Initially, We use the results from the 3B (0.5) prompts as the baseline for multi-prompt segmentation in SAM, MedSAM, and MPA-SAM on the standard medical imaging datasets, specifically shown in the 3$^{rd}$ row, 7$^{th}$ row, and 11$^{th}$ row.

\begin{table}[htbp]
    
    
    \centering
    \caption{Comparison of model performance under uncertainty-guided prompt adaptation with BBox overlap ratio of 0.5. Here, UGPA$^1$, UGPA$^2$, UGPA$^3$, UGPA$^4$ represent refined 3B, refined 3P \& 3B, refined 5P \& 3B, refined 10P \& 3B. The top-2 results are highlighted in \textbf{bold} and \underline{underline}.}
    \setlength{\tabcolsep}{4pt}

     \label{tab:uncertainty-guided-multi-prompt_iou0.5}
     \begin{tabular}{lcccccccc}
    
        \toprule
 
        \multirow{2}{*}{\textbf{Modality}} &
            \multicolumn{2}{c}{\textbf{Dermoscopy}}&
            \multicolumn{2}{c}{\textbf{Colonoscopy}}&
            \multicolumn{2}{c}{\textbf{Ultrasound}}&
            \multicolumn{2}{c}{\textbf{Avg}} \\
            \cmidrule(lr){2-3} \cmidrule(lr){4-5} \cmidrule(lr){6-7} \cmidrule(lr){8-9} 
            & {IoU} & {Dice} & {IoU} & {Dice} & {IoU}  & {Dice} & {IoU} & {Dice}
             \\
            \midrule
        \centering
        \multirow{1}{*}{\textbf{w/o UGPA}}  & 0.619	& 0.750	& 0.559	& 0.698	& 0.673	& 0.796	& 0.617	& 0.748	
    \\ 
        \centering
        \multirow{1}{*}{\textbf{UGPA$^1$}} & 
            {0.774} & {0.863}  &
            {0.751} & {0.843} &
             {0.889} &  {0.939}  & {0.805}	& {0.882}
             \\
        \centering
        \multirow{1}{*}{\textbf{UGPA$^2$}} &  0.779	& 0.867	& 0.755	& 0.846	& 0.891	& 0.941	& 0.808	& 0.885	
 \\ 
        \centering
        \multirow{1}{*}{\textbf{UGPA$^3$}} & \underline{0.796} & \underline{0.880}  & \textbf{0.784} & \textbf{0.871} &  \textbf{0.898} & \underline{0.945}  & \textbf{0.826} & \textbf{0.899} \\
        \centering
        \multirow{1}{*}{\textbf{UGPA$^4$}} & 
            \textbf{0.801} & \textbf{0.883} &
            \underline{0.779} & \underline{0.867}  &
            \textbf{0.898} & \textbf{0.946} &
            \textbf{0.826} & \textbf{0.899} \\ \bottomrule
 
\end{tabular}
\end{table}

\begin{table}[htbp]
    
    
    \centering
    \caption{Comparison of model performance under uncertainty-guided prompt adaptation with BBox overlap ratio of 0.75. Here, UGPA$^1$, UGPA$^2$, UGPA$^3$, UGPA$^4$ represent refined 3B, refined 3P \& 3B, refined 5P \& 3B, refined 10P \& 3B. The top-2 results are highlighted in \textbf{bold} and \underline{underline}.}
    \setlength{\tabcolsep}{4pt}
     \label{tab:uncertainty-guided-multi-prompt_iou0.75}
    \begin{tabular}{lcccccccc}
    
        \toprule
 
        \multirow{2}{*}{\textbf{Modality}} &
            \multicolumn{2}{c}{\textbf{Dermoscopy}}&
            \multicolumn{2}{c}{\textbf{Colonoscopy}}&
            \multicolumn{2}{c}{\textbf{Ultrasound}}&
            \multicolumn{2}{c}{\textbf{Avg}} \\
            \cmidrule(lr){2-3} \cmidrule(lr){4-5} \cmidrule(lr){6-7} \cmidrule(lr){8-9} 
            & {IoU} & {Dice} & {IoU} & {Dice} & {IoU}  & {Dice} & {IoU} & {Dice}
             \\
            \midrule
        \multirow{1}{*}{\textbf{w/o UGPA}} 	& 0.792	& 0.877	& 0.766	& 0.859	& 0.885	& 0.938	& 0.814	& 0.892
    \\ 
        \centering
        \multirow{1}{*}{\textbf{UGPA$^1$}} & 
             {0.815} & {0.892} &
             {0.800} & {0.880} &
               {0.909} &  {0.952} & {0.842}	& {0.908}
             \\
        \centering
        \multirow{1}{*}{\textbf{UGPA$^2$}} & 0.817	& 0.893	& 0.805	& 0.885	& 0.911	& 0.953	& 0.844	& 0.910
 \\ 
        \centering
        \multirow{1}{*}{\textbf{UGPA$^3$}} & \underline{0.836} & \underline{0.906}  & \underline{0.832} & \underline{0.902} &  \textbf{0.920} & \textbf{0.958}  & \underline{0.862} & \underline{0.922} \\
        \centering
        \multirow{1}{*}{\textbf{UGPA$^4$}} & 
            \textbf{0.839} & \textbf{0.909} &
             \textbf{0.833} & \textbf{0.903} &
              \textbf{0.920} & \textbf{0.958} &
            \textbf{0.864} & \textbf{0.923} \\ \bottomrule
 
\end{tabular}
\end{table}

By analyzing the results from the 3$^{rd}$ row to 4$^{th}$ row for SAM and from the 7$^{th}$ row to 8$^{th}$ row for MedSAM, it is evident that the combination of point and box prompts leads to a decrease in Dice scores from 52.6\% to 44.8\% for SAM, and from 63.3\% to 37.7\% for MedSAM. The analysis of the table reveals that the performance of SAM and MedSAM does not exhibit improvement with the incorporation of multi-type prompts. Moreover, the use of point prompts may negatively impact the effectiveness of box prompts, resulting in a deterioration of segmentation performance. Furthermore, with the introduction of MPA, we observed that the performance of MPA-MedSAM improved by approximately 3.5\% between 11$^{th}$ row and 12$^{th}$ row, further demonstrating the effectiveness of combining point prompts with box prompts was substantiated.

\subsubsection{Effectiveness of Uncertainty-Guided Prompts Adaptation}
 We conducted a comprehensive ablation study to validate the effectiveness of the proposed UGPA. The results are presented in Table~\ref{tab:uncertainty-guided-multi-prompt_iou0.5} and Table~\ref{tab:uncertainty-guided-multi-prompt_iou0.75}, where the baseline
 (1$^{st}$ row) represents 3 Bboxes from low-quality Bbox random shift provided by the user, serving as the initial input for our proposed MedSAM-U, without UGPA. 
As shown in the 2$^{nd}$ row and 3$^{rd}$ row in the Table~\ref{tab:uncertainty-guided-multi-prompt_iou0.5} and Table~\ref{tab:uncertainty-guided-multi-prompt_iou0.75}, when combining box prompts with point prompts, there is an improvement compared to using the initial 3 Bboxes without UGPA. The combined approach also demonstrates enhanced performance relative to refining Bboxes alone, both in the 3B (0.5) and 3B (0.75) settings. 
Additionally, in the Table~\ref{tab:uncertainty-guided-multi-prompt_iou0.5} and Table~\ref{tab:uncertainty-guided-multi-prompt_iou0.75}, from the 3$^{rd}$ row to the 5$^{th}$ row, we observe 1.3\% improvement for the 3B (0.5) setting and a 1.3\% improvement for the 3B
(0.75) setting with the increase in the number of points. 
Specifically, increasing the points increasing from 5 to 10 points, results are in minimal enhancement. It does not mean that the more points there are, the better the effect will be. In fact, the improvement in effect may tend to saturation, or even counter-effectively in some cases.  

The segmentation results of our method in different modalities are shown in the Fig.~\ref{fig:F_4}, three Bboxes were generated by introducing variations to GT box in terms of position and shape to simulate user input, serving as the input for our proposed MedSAM-U method. Under the refinement provided by MedSAM-U, the results showed significant improvement. The method effectively leveraged the uncertainty map, enhancing the accuracy and robustness of the segmentation.

\section{Conclusion}
In this study, we introduced MedSAM-U, a novel model designed with an uncertainty-guided, auto-refining multi-prompt approach for reliable and accurate medical image segmentation. First, MPA-MedSAM was utilized to adapt various multi-prompt strategies for MedSAM. We then implemented UGMP to estimate uncertainty in the segmentation results without adding additional training parameters. Crucially, we developed a novel uncertainty-guided multi-prompt adaptation method that automatically generates reliable prompts, leading to highly accurate segmentation results. Furthermore, by training on multi-datasets from multiple modalities, MedSAM-U effectively functions as a universal image segmentation model. Experimental results across five distinct modal datasets show that within the BBox overlap ratio range of 0.5 to 0.75, MedSAM-U significantly improved performance, with average improvements ranging from 1.7\% to 20.5\%, compared to the baseline MedSAM model. Additionally, our results indicated that the lower the initial BBox quality, the greater the improvement achieved by MedSAM-U.

Moving forward, our research will concentrate on two key areas. First, we plan to directly estimate uncertainty within the adapter. Second, we aim to harness this uncertainty to achieve automantic advanced annotation, enabling AI and the model to perform automatic annotations without human intervention.

\bibliographystyle{IEEEtran}
\bibliography{Refer}
\end{document}